\newif\ifarxiv

\ifarxiv
  \documentclass[conference]{IEEEtran} %
\else
  \documentclass[letterpaper,10pt,conference]{ieeeconf} %
  \IEEEoverridecommandlockouts    %
\fi

\usepackage{amssymb}
\usepackage{amsmath}
\usepackage{graphicx} 
\usepackage{subcaption}
\usepackage{booktabs}
\usepackage{wrapfig}
\usepackage{multirow}
\usepackage{url}
\usepackage{xcolor}

\newcommand{\methodname}[1]{\textsc{MimicDroid}}

\IEEEoverridecommandlockouts                              %

\usepackage[backend=biber,
            hyperref=true,
            url=false,
            isbn=false,
            doi=false,
            backref=false,
            style=ieee,
            citestyle=numeric-comp,
            sorting=none,%
            block=none]{biblatex}
\AtEveryBibitem{%
  \clearfield{pages}%
  \clearfield{volume}%
  \clearfield{number}%
  \clearfield{issue}%
  \clearfield{month}%
}

\DeclareNameFormat{author}{%
  \nameparts{#1}%
  \usebibmacro{name:family}{\namepartfamily}{\namepartgiven}{\namepartprefix}{\namepartsuffix}%
  \usebibmacro{name:andothers}}
\DeclareNameFormat{editor}{%
  \nameparts{#1}%
  \usebibmacro{name:family}{\namepartfamily}{\namepartgiven}{\namepartprefix}{\namepartsuffix}%
  \usebibmacro{name:andothers}}

\title{\LARGE \bfseries \methodname{}: In-Context Learning for Humanoid Robot Manipulation\\
from Human Play Videos}

\author{Rutav Shah$^1$\quad Shuijing Liu$^{*,1}$\quad Qi Wang$^{*,1}$\quad Zhenyu Jiang$^{*,1}$\quad Sateesh Kumar$^1$\quad Mingyo Seo$^1$\\ Roberto Mart{\'i}n-Mart{\'i}n$^{1,2}$\quad Yuke Zhu$^{1,3}$%
\\
$^1$The University of Texas at Austin\quad $^2$Amazon Consumer Robotics\quad $^3$NVIDIA%
\thanks{$^{*}$ Equal contribution. Correspondence: rutavms@utexas.edu}%
}

\begin{document}
\twocolumn[{%
    \renewcommand\twocolumn[1][]{#1}%
    \maketitle
    \vspace*{-5mm}

    \begin{center}
        \centering
        \includegraphics[width=\textwidth, trim={0 220 322 0}, clip]{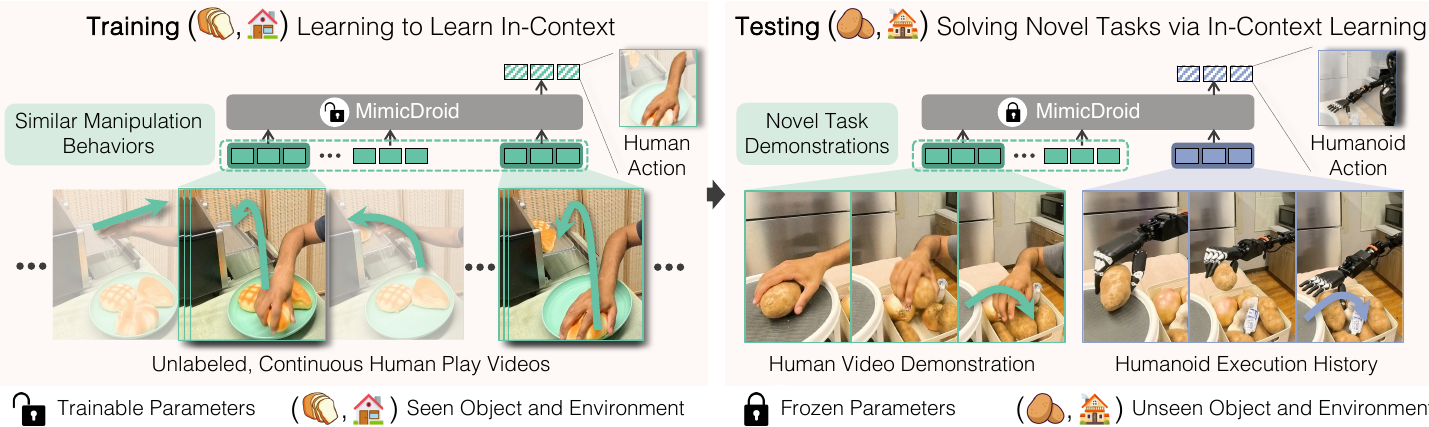}
        \captionof{figure}{
        \textbf{Overview.}~\methodname{} enables few-shot learning for humanoid manipulation by training solely on human play videos---a scalable and diverse data source.
        At test time, it observes human videos of novel tasks and uses in-context learning to perform the same tasks under different object placements.
        }
        \label{fig:pull_figure}
    \end{center}%
}]

\thispagestyle{empty}
\pagestyle{empty}
\begingroup
  \renewcommand\thefootnote{\fnsymbol{footnote}}
  \setcounter{footnote}{1} %
  \footnotetext{Equal contribution. Correspondence: rutavms@utexas.edu}
\endgroup
\begin{abstract}
We aim to enable humanoid robots to efficiently solve new manipulation tasks from a few video examples.
In-context learning (ICL) is a promising framework for achieving this goal due to its test-time data efficiency and rapid adaptability.
However, current ICL methods rely on labor-intensive teleoperated data for training, which restricts scalability.
We propose using human play videos---continuous, unlabeled videos of people interacting freely with their environment---as a scalable and diverse training data source.
We introduce \methodname{}, which enables humanoids to perform ICL using human play videos as the only training data.
\methodname{} extracts trajectory pairs with similar manipulation behaviors and trains the policy to predict the actions of one trajectory conditioned on the other.
Through this process, the model acquired ICL capabilities for adapting to novel objects and environments at test time.
To bridge the embodiment gap, \methodname{} first retargets human wrist poses estimated from RGB videos to the humanoid, leveraging kinematic similarity.
It also applies random patch masking during training to reduce overfitting to human-specific cues and improve robustness to visual differences.
To evaluate few-shot learning for humanoids, we introduce an open-source simulation benchmark with increasing levels of generalization difficulty.
~\methodname{} outperformed state-of-the-art methods and achieved nearly twofold higher success rates in the real world.
Additional materials can be found on: \url{ut-austin-rpl.github.io/MimicDroid}
\end{abstract}

\section{Introduction}
Humanoid robots are well-suited for diverse household manipulation tasks due to their human-like morphology.
Yet homes exhibit large variability: objects, layouts, and tasks change across time and households.
To effectively cope with such variability, humanoids must move beyond pre-defined sets of behaviors and adapt rapidly to novel situations.
For instance, when a new appliance is installed, the robot should be able to acquire the necessary skills to manipulate it from a handful of demonstrations, a problem setting known as few-shot learning~\cite{MAML, DAML, james2018task}.

In-context learning (ICL) has shown promise for few-shot learning, offering data-efficient and rapid adaptation at test time~\cite{gpt3,flamingo,xu2022prompting,amago,algodistil,regent,kat,icrt,instantpolicy}.
By simply conditioning on a few human demonstrations, ICL can predict robot actions to achieve novel tasks at test time without expensive retraining (Fig.~\ref{fig:pull_figure}, Right).
However, effective ICL for few-shot learning relies on large and diverse training data~\cite{raparthy2023generalization,wang2024benchmarking,kirsch2022general}.
In manipulation, prior ICL methods rely on teleoperated robot demonstrations as training data~\cite{icrt,ricl}, which are expensive and time-consuming to scale~\cite{roboturk,rtx,droid}.
This limitation motivates us to explore scalable training data sources and methods that can leverage such data to enable ICL for humanoid manipulation.

A promising alternative training data source is human play videos---continuous recordings of people interacting with their environments, typically spanning $10$--$20$ minutes of interaction and driven by their curiosity.
Human play videos capture task-agnostic, unscripted, and exploratory interactions with environments~\cite{mimicplay,latent-play}.
Compared to teleoperated demonstrations, they are approximately $18\times$ faster to collect~\cite{mimicplay} and inherently diverse, covering a broad range of tasks, object configurations, and manipulation behaviors~\cite{latent-play}.
Leveraging these advantages, we explore scalable and diverse human play videos to train ICL policies that perform few-shot learning for humanoid manipulation.

To realize the potential of human play videos for ICL, two key challenges must be addressed. 
First, the model should learn tasks from in-context examples, an ability that meta-training can instill through \textit{learning to learn} in-context~\cite{gpt3,metaicl}.
However, this requires constructing meta-training samples from raw human play videos in a scalable, self-supervised way.
Second, the kinematic and visual gap between human and robot embodiments presents challenges for applying ICL at test time.
The kinematic gap must be bridged so actions learned from human videos can be executed on the robot without losing task intent.
The visual gap must be addressed to avoid overfitting to human appearances, which can hinder its ability to apply ICL on robots.

To address these challenges, we develop~\methodname{} (a \textbf{Mimic}king an\textbf{Droid}), a novel method to perform few-shot humanoid manipulation via ICL using only RGB human play videos for training.
At test time, \methodname{} is provided with $1$--$3$ videos of a human performing a task, potentially involving novel objects and environments, and applies ICL to mimic and perform the task (Fig.~\ref{fig:pull_figure}).
Effective ICL relies on exploiting recurring patterns in observation-action relationships, i.e., how visual observations correspond to the actions that follow, enabling the model to predict actions in new scenes.
To instill this capability, we generate meta-training samples by pairing trajectory segments with similar patterns, treating one as the target and the others as proxies for test-time demonstrations.
This construction encourages the policy to exploit similarities in observation-action relationships across trajectories to predict actions.
Human play videos are well-suited for this construction, as they naturally exhibit recurring patterns of similar manipulation behaviors, like ``moving \textit{bread} from plate to oven'' and later ``moving \textit{bagel} from plate to oven,'' which \methodname{} retrieves to construct meta-training samples in a self-supervised manner.
To bridge the kinematic gap, \methodname{} retargets predicted human wrist poses to humanoid wrist poses at test time.
By operating in task space (Cartesian wrist pose) and exploiting kinematic similarities between embodiments, it preserves the underlying task intent~\cite{okami,ph2d,mistry2015representation}.
To mitigate the visual gap, it applies random patch masking during training~\cite{mae,crossmae}, reducing overfitting to human-specific appearances.

We evaluate \methodname{} in both simulation and real-world settings.
We build a new simulation benchmark for evaluating few-shot learning in humanoid manipulation spanning three generalization levels (Sec.~\ref{sec:exp_setup_main}).
By exploiting observation-action relationships in human videos through ICL, \methodname{} outperforms task-conditioned baselines~\cite{vid2robot,h2r}, achieving a twofold improvement in real-world success rate.
Compared to parameter-efficient fine-tuning~\cite{hand}, ICL adapts instantaneously and achieves a $26\%$ higher success rate at test time.
We show that~\methodname{} scales effectively with training data, yielding a gain of $20\%$ when increasing training human play videos from $64$k to $320$k frames.
Our main contributions are as follows.
\begin{enumerate}
    \item \methodname{} enables few-shot learning for humanoid manipulation via ICL using only human play videos.  
    \item We introduce a new simulation benchmark~\cite{robocasa,robosuite} with $8$ hours of play data to evaluate few-shot learning for humanoids.  
    \item \methodname{} outperforms prior works and demonstrates scalability with data, while our analysis highlights current limitations and future directions.  
\end{enumerate}

\begin{figure*}[th]
    \centering
    \includegraphics[trim={0px 215px 380px 0}, width=0.87\textwidth]{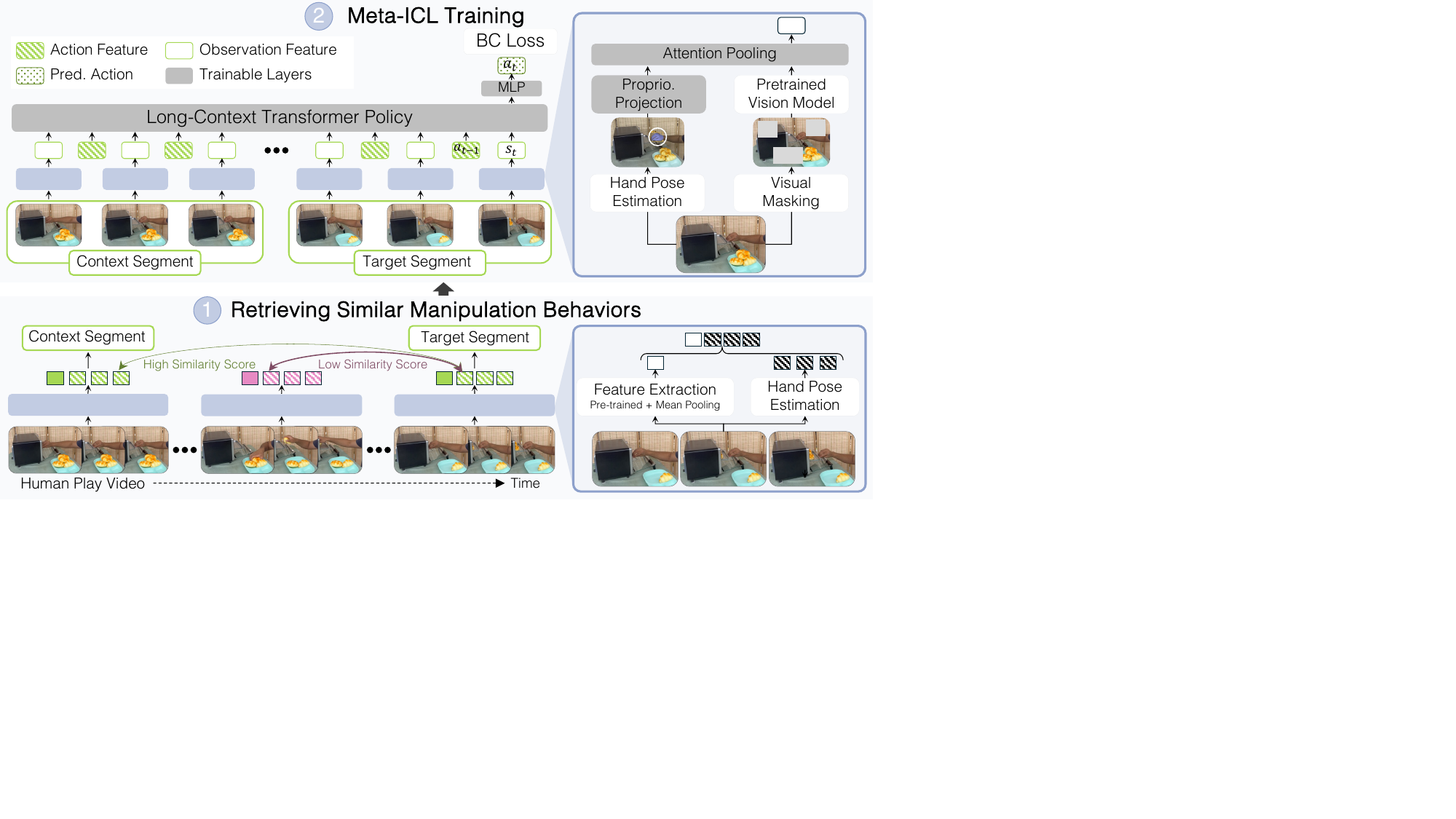}
    \caption{
    \textbf{Method Overview.} \methodname{} performs meta-training for in-context learning (Meta-ICL) by constructing context-target pairs from human play videos.
    For a target segment, we retrieve the top-$k$ most similar trajectory segments (bottom-left) based on observation-action similarity (bottom-right) to serve as context.
    These context-target pairs are used to teach the policy in-context learning (top-left).
    To overcome the human-robot visual gap and avoid overfitting to human-specific visual cues, we apply visual masking to input images (top-right), improving transferability.
    }
    \vspace{-0pt}
    \label{fig:method}
\end{figure*}
\section{Related Work}
\textbf{Few-Shot Learning in Manipulation.}
Few-shot learning is the ability to learn new tasks from just a few demonstrations, enabling robots to adapt to novel tasks and scenes.
Meta-learning~\cite{schmidhuber1987evolutionary,naik1992meta,santoro2016meta,hochreiter2001learning} provides one approach to such adaptation, where the model acquires a learning strategy that allows it to quickly learn new skills at test time with only a few examples.
Early approaches in robot manipulation instantiated this idea through gradient-based meta-learning methods~\cite{MAML,DAML}.
While data-efficient, these methods lack on-the-fly adaptation.
Recent progress in ICL~\cite{duan2016rl, gpt3, flamingo}, driven by long-context transformers~\cite{vaswani2017attention} and meta-training on diverse data, allows models to adapt on-the-fly by matching patterns in the input-output examples~\cite{duan2016rl,gpt3,flamingo}.
Meta-training for in-context learning (Meta-ICL) amplifies this ability by training on structured context-target pairs~\cite{metaicl}.
In robotics, Meta-ICL has shown preliminary promise in learning visuomotor policies with robot teleoperation data or simulated data~\cite{icrt,instantpolicy,ricl}. However, since the effectiveness of ICL heavily depends on large and diverse training datasets~\cite{raparthy2023generalization,wang2024benchmarking,kirsch2022general}, these methods are limited by the high cost of data collection~\cite{icrt} and the narrow diversity of simulated tasks~\cite{instantpolicy}.

\textbf{Learning from Human Videos.}
Human videos offer a more scalable and diverse data source for robot learning, compared to expensive and time-consuming robot teleoperation data. Prior work has explored various strategies to extract knowledge from human videos, such as learning visual representations~\cite{shah2021rrl, nair2022r3m, majumdar2023we} and deriving reward functions~\cite{shao2021concept2robot, ma2023liv, ma2022vip, hudor}.
However, these approaches often require additional robot data in addition to human videos to accomplish the downstream task~\cite{bahety2024screwmimic}.
Another line of work focuses on extracting motion priors directly from human videos~\cite{ph2d, kareer2024egomimic, mimicplay}, but often depends on specialized hardware such as VR or hand-tracking devices, which add overhead and limit scalability.
To overcome this, recent works leverage advances in hand pose estimation models~\cite{wilor,hamer} to extract action information directly from RGB videos~\cite{lepert2025phantom, okami, orion}.
Among these works, some approaches~\cite{lepert2025phantom, orion} require human demonstrators to mimic the morphology and joint constraints of specific robot manipulators with unnatural motions.
Humanoid robots reduce the necessity for such unnatural motions by leveraging their kinematic similarity to humans, which enables a natural mapping from human motions to robot actions~\cite{okami,ph2d}.

In summary, to solve few-shot learning for humanoid manipulation, \methodname{} builds on two pillars:
(1) ICL to allow the humanoid to adapt to novel objects and environments on the fly;
(2) human play videos, consisting of raw RGB frames, as a scalable and diverse training data source to build ICL foundations by leveraging the success of hand pose estimation and the kinematic similarities between humans and humanoids.

\section{Methodology}
In this section, we present an overview (Sec.~\ref{sec:method:overview}), components involved in constructing the training data (Sec.~\ref{sec:method:training_sample}), and overcoming the embodiment gap between human and robot (Sec.~\ref{sec:method:embodiment}), and finally, meta-training objective (Sec.~\ref{sec:method:training_objective}) to enable test-time ICL (Fig.~\ref{fig:method}).
\subsection{Overview}\label{sec:method:overview}
\textbf{Problem Setup.} 
Robots deployed in real-world environments must handle diverse objects, layouts, and environments, making it infeasible to predefine all possible task variations.  
To enable generalization in such settings, we consider a test-time scenario where the robot is given a small set of human demonstration trajectories for a novel task $\mathcal{T} \sim p(\mathcal{T}_{\text{test}})$: $
\mathcal{D}_{\text{test}} = \{\tau_i^{\text{demo}}\}_{i=1}^k,
$
where each $\tau_i^{\text{demo}} = \{s_t\}_{t=1}^{T_i}$ is a sequence of RGB frames of a human performing $\mathcal{T}$.  
The goal is to learn a visuomotor policy $\pi_\theta$ that can leverage these demonstrations to perform the task.

\textbf{Training Data.} 
During training, \methodname{} is provided with a dataset 
$
\mathcal{D}_{\text{train}} = \{\tau_i\}_{i=1}^N
$
consisting of human play trajectories. 
Each trajectory $\tau_i$ is a sequence of RGB frames from a single play session, where $T_i$ denotes the trajectory length, $\quad \tau_i = \{s_t\}_{t=1}^{T_i}$, typically corresponding to $10$--$20$ minutes of interaction. 
They span a diverse set of tasks, forming an implicit task distribution $\mathcal{T}_{\text{train}}$.

\textbf{Policy Learning}
Using $\mathcal{D}_{\text{train}}$, we aim to train a visuomotor policy $\pi_\theta$ that can perform in-context learning (ICL). We cast this training into the Meta-ICL framework~\cite{metaicl}, where a task $\mathcal{T}$ is sampled from $p(\mathcal{T}_{\text{train}})$, $\sigma_{\mathcal{T}}^{\text{ctx}}$ denotes a set of \textit{context} trajectories, and $\sigma_{\mathcal{T}}^{\text{tgt}}$ is the \textit{target} trajectory. The context trajectories provide examples of how the task is performed, while the target trajectory supervises the policy's action predictions conditioned on the context. We use behavior cloning to supervise the policy by minimizing the loss $\mathcal{L}_{\mathcal{T}}$ between the predicted and ground-truth actions on $\sigma_{\mathcal{T}}^{\text{tgt}}$. Through this process, $\pi_\theta$ learns to exploit recurring observation-action patterns in context trajectories to predict actions for the target trajectory.
Formally, the training objective is
\begin{equation}
\label{eq:metaobj}
\mathbb{E}_{\mathcal{T} \sim p(\mathcal{T}_{\text{train}})} \;
\mathbb{E}_{\sigma_{\mathcal{T}}^{\text{ctx}}, \, \sigma_{\mathcal{T}}^{\text{tgt}} \sim \mathcal{T}}
\left[ \mathcal{L}_{\mathcal{T}} \big( \pi_\theta(\sigma_{\mathcal{T}}^{\text{tgt}} \mid \sigma_{\mathcal{T}}^{\text{ctx}}) \big) \right],
\end{equation}

Unlike traditional meta-learning, which depends on an explicitly defined discrete set of tasks in $p(\mathcal{T}_{\text{train}})$, our setting removes this assumption and instead relies on an implicit task distribution induced by natural human interactions during play. 
Moreover, play videos lack task labels to sample context-target trajectories for a given task $\sigma_{\mathcal{T}}^{\text{tgt}}, \sigma_{\mathcal{T}}^{\text{ctx}} \sim \mathcal{T}$, and explicit action information to supervise the policy $\mathcal{L}_{\mathcal{T}}$.

\textbf{Evaluation Protocol.} 
Given $\mathcal{D}_{\text{test}}$, and conditioned on past observations $s_{1:t}$ and past actions $a_{1:t-1}$, the trained policy $\pi_\theta$ predicts the next action $a_t$ to successfully perform the task. 
The policy uses these demonstrations for in-context learning (ICL), adapting on the fly \textit{without additional parameter updates}. 
For systematic evaluation, we assess generalization across three test-time task distributions with increasing difficulty introduced via novel objects and environments.

\textbf{Challenges.} To enable effective test-time ICL capabilities, the main challenges include:
(a) \textit{Constructing training samples.}
Human play videos provide only RGB frames $\tau = \{s_t\}_{t=1}^{T}$, whereas training requires state-action pairs $\tau = \{(s_t, a_t)\}_{t=1}^{T}$ for supervision, along with proprioceptive signals for richer input. Thus, the missing actions and proprioception should be estimated.
A subsequent challenge lies in constructing training samples consisting of context-target pairs from long, continuous human play videos for Meta-ICL (See Eq.~\ref{eq:metaobj}) (Sec.~\ref{sec:method:training_sample});
(b) \textit{Overcoming the embodiment gap.}
Kinematic and visual differences between humans and the humanoid make it challenging to transfer learned ICL capabilities at test time (Sec.~\ref{sec:method:embodiment}).
\subsection{Constructing Training Samples }\label{sec:method:training_sample}
\textbf{Extracting low-level action and proprioceptive information.}
To fill in the missing action information, we use the future human hand pose $k$ timesteps ahead as the intended action for the current timestep.
~\methodname{} estimates the hand poses from human RGB frames using an off-the-shelf hand pose estimation model $f$~\cite{wilor}, leveraging recent advances in vision-based hand tracking.
This avoids overhead costs introduced by using specialized hardware to predict hand pose~\cite{mimicplay,ph2d,kareer2024egomimic}.
Specifically, given an RGB frame, the model $f$ predicts the wrist pose and a grasp signal derived from finger joint angles.
The predicted hand pose $h_{t+k} = f(s_{t+k})$ is treated as the action label $a_t$.
In addition, the hand pose at each time step $h_t$ is included in the observation, serving as the proprioception, to provide richer input for learning.
This transforms a raw trajectory from the play data $\tau$ into a processed trajectory with inferred low-level actions and proprioceptives, $\tau' = \{s'_t, a_t\}_{t=1}^{T}$, where $s'_t = \{s_t, h_t\}$.
\begin{figure}[ht]
  \centering
  \vspace{-5pt}
  \includegraphics[trim={0px 5px 60px 0}, width=\linewidth]{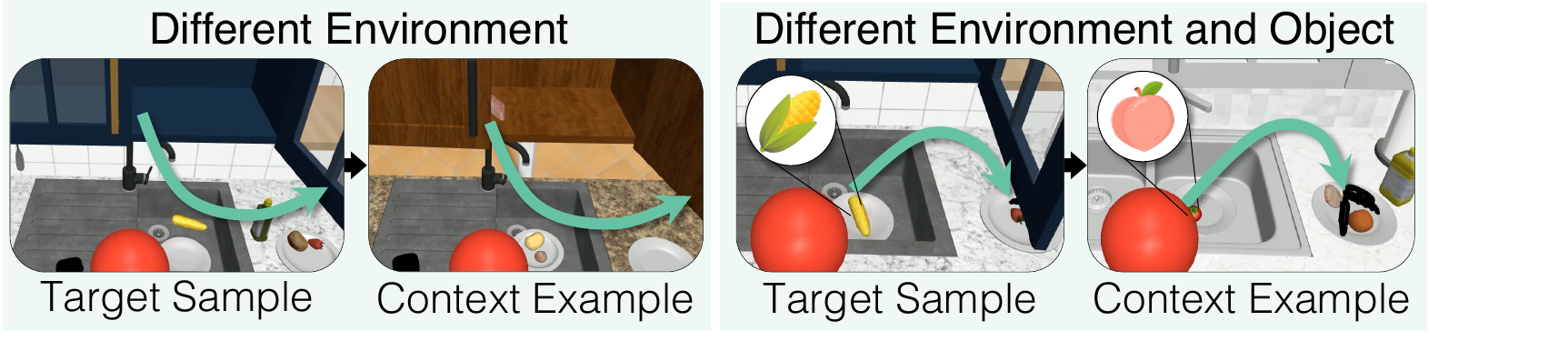}
  \vspace{-15pt}
  \caption{Examples of target and retrieved context examples.}
  \vspace{-5pt}
  \label{fig:mew_similarity}
\end{figure}

\textbf{Constructing context-target training pairs.}
At test time, the policy must perform novel tasks, \textit{e.g.}, using a new appliance by mimicking human demonstrations.
This requires leveraging observation-action relationships and their recurring patterns across demonstrations to predict actions.
To enable this, we construct Meta-ICL training samples by pairing trajectory segments with similar patterns, designating one as the target and the others as context (Eq.~\ref{eq:metaobj}).
Human play videos are well-suited for constructing Meta-ICL training data because they naturally exhibit such patterns, \textit{e.g.}, moving an item from one receptacle to another and later moving a different item to another receptacle, each following similar observation-action relationships (Fig.~\ref{fig:mew_similarity}).

To retrieve such similar patterns from play videos, specifically, we randomly sample a target trajectory segment $\sigma^{\text{tgt}} \subset \tau'$, and then identify the top $k$ most similar segments $\{\sigma_{i}^{ctx}\}_{i=1}^{k}$ from the dataset to serve as context, where $k$ is a hyperparameter.
The similarity between two trajectory segments $\sigma_{x}$ and $\sigma_{y}$ is computed as the cosine similarity between their feature embeddings,
$d(\sigma_{x}, \sigma_{y}) = \cos\!\left(\phi(\sigma_{x}), \phi(\sigma_{y})\right)$, 
where $\phi(\cdot)$ denotes the feature embedding function.
To make segment features $\phi(\sigma)$ robust to visual noise (\textit{e.g.}, background clutter) and capture the observation-action distribution, we first extract frame-wise visual features using the pretrained vision model $g$~\cite{dinov2}. We then apply temporal mean pooling over the features, and concatenate the result with the sequence of actions to form the final segment features:
\begin{align}
\phi(\sigma) = \Big[\frac{1}{T}\sum_{t=1}^{T}g(s_t),\ a_1,\ \ldots,\ a_T\Big]
\end{align}
In summary, \methodname{} leverages the inherent repetitive manipulation behaviors naturally present in human play data to generate training samples for Meta-ICL (Fig.~\ref{fig:method}, bottom).

\subsection{Overcoming the Embodiment Gap}\label{sec:method:embodiment}
To transfer policies from human play videos to humanoid robots, \methodname{} must overcome embodiment gaps between the two.
To bridge the kinematic gap, it retargets predicted human wrist poses to humanoid wrist poses and applies inverse kinematics to compute joint angles. By operating in task space and exploiting kinematic similarities, it preserves task intent while avoiding the need for demonstrators to mimic robot morphology~\cite{okami,ph2d}.
Moreover, differences in body appearance and occlusion patterns between humans and humanoids introduce a visual gap that can hinder ICL at test time.
To reduce overfitting to the human-specific visual cues,~\methodname{} incorporates visual masking during training.
Specifically, during training, random patches between $1$ and $n$ are masked in the input images. 
This operation is applied with a probability $p$, encouraging the model to rely less on superficial human-specific cues and instead learn representations that generalize across embodiments (Fig.~\ref{fig:method}, top-right).
As a result,~\methodname{} can learn a policy using \textit{only} human videos without \textit{any} teleoperated demonstrations and deploy it on the robot.

\begin{figure*}[ht!]
    \centering
    \includegraphics[trim={0 340px 400px 0}, width=0.93\linewidth]{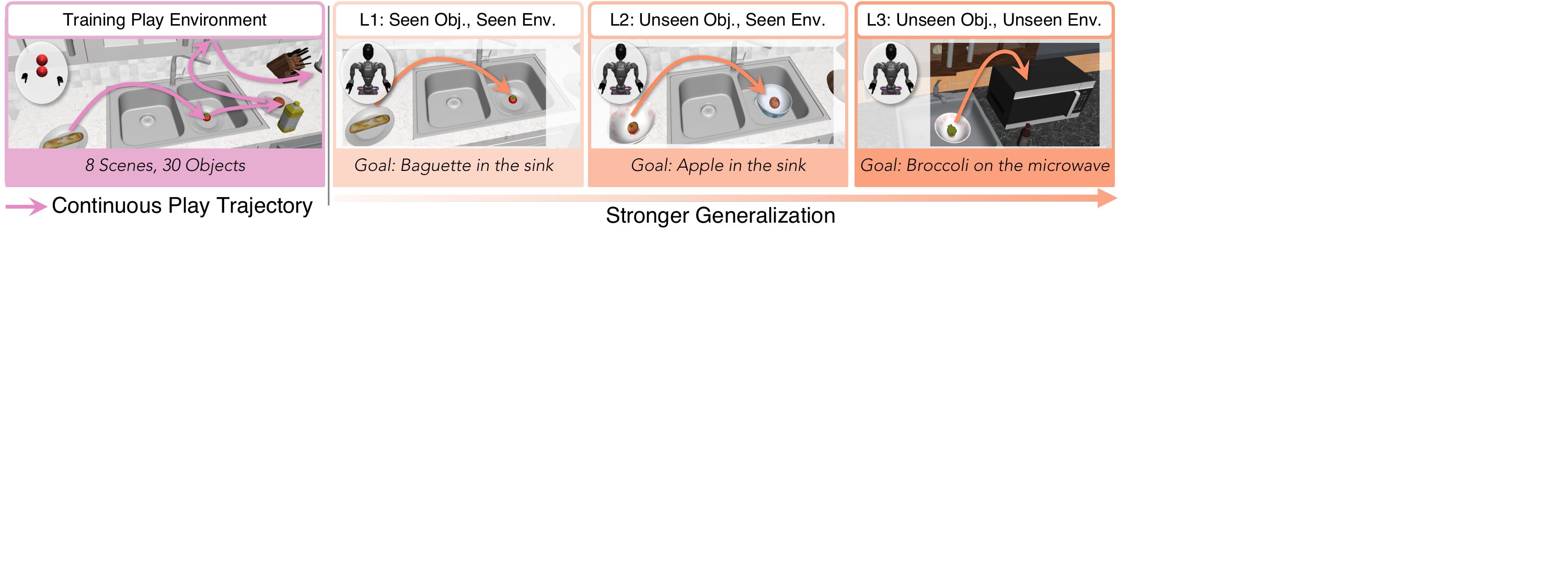}
    \caption{
    \textbf{Overview of our simulation benchmark.} We introduce a simulation benchmark to evaluate few-shot learning for humanoid manipulation.  
    It contains $8$ hours of play data collected using free-floating hands (Left) across $30$ objects and $8$ kitchen environments.  
    Evaluation is structured into three levels: L1, L2, and L3 with increasing difficulty via novel objects and environments, enabling systematic assessment of generalization to humanoids (target) (Right).  
    The bottom-right of each image indicates the source and target embodiments. Language descriptions are included for clarity.
    }
    \vspace{-10pt}
    \label{fig:sim_benchmark}
\end{figure*}

\subsection{Meta-training for In-Context Learning}\label{sec:method:training_objective}
To train the policy to perform ICL, we optimize the meta-learning objective in Eq.~\ref{eq:metaobj} using behavior cloning.
Instead of assuming an explicitly defined discrete set of tasks in $p(\mathcal{T}_{\text{train}})$, we rely on an implicit task distribution induced by continuous human play.
This is advantageous because human play naturally spans diverse manipulation behaviors, providing a richer and more scalable source of task variation than a manually defined set.
We approximate the implicit distribution by uniformly sampling target trajectories from play videos, with each trajectory serving as a task instance.
This exposes the policy to a wide variety of behaviors, while context trajectories for training are retrieved as described in Sec.~\ref{sec:method:training_sample}.
This meta-training process exposes the policy to many such context-target pairs, enabling it to learn a general strategy for adapting to new tasks at test time.

Formally, each training instance consists of $k$ context trajectory segments $\{\sigma^{ctx}_{i}\}_{i=1}^{k}$ and a target segment $\sigma^{\text{tgt}}$ where the target segment is $\sigma^{\text{tgt}} = \{(s'_t, a_t)\}_{t=1}^{T}$ with $s'_t$ denotes the RGB images with proprioceptive information, and $a_t$ is the target action extracted from future hand poses (Sec.~\ref{sec:method:training_sample}).
To model the multimodal nature of human play data, we adopt action chunking, where the policy predicts a sequence of $l$ future actions at each time step instead of only the immediate next action~\cite{act_chunking}.
The policy $\pi$ is trained to imitate the extracted actions from the target segment by minimizing the L1 loss between predicted and ground-truth action chunks (defined for one sample):
\begin{equation}
    \mathcal{L}_{\text{BC}} = \sum_{t=1}^{T-1} \left\| \pi\Big(a_{t:t+l}^{\text{pred}}\ \Big|\ s'_{1:t},a_{1:t-1},\ \{\sigma_{c,i}\}_{i=1}^{k}\Big) - a_{t:t+l} \right\|_1
\end{equation}
Here, $a_{t:t+l}$ denotes the ground-truth $l$-step action sequence from time $t$, and $\pi(.)$ denotes the outputs corresponding $l$-step action prediction conditioned on the past trajectory and the $k$ context examples.

\section{Experiment Setup}\label{sec:exp_setup_main}
\textbf{Simulation Benchmark.}
Until now, there has been no standardized benchmark to systematically evaluate few-shot learning in humanoid manipulation policies. To fill this gap and facilitate future research, we introduce a novel simulation benchmark with a wide range of objects, tasks, and environments, building on RoboCasa~\cite{robocasa2024}.
The benchmark includes humanoid manipulation tasks, such as pick-and-place tasks involving various objects and receptacles, as well as the manipulation of articulated objects like faucets and cabinets.
We systematically categorize the tasks into progressively harder generalization levels (Fig.~\ref{fig:sim_benchmark}):\\
\textit{L1 (Seen Objects, Seen Environment)}:
The robot must perform manipulation tasks with objects it encountered during training in the environments. This level assesses the model's ability to generalize to new object positions.
\\
\textit{L2 (Unseen Objects, Seen Environment)}: 
The task requires the robot to apply the learned manipulation skills to novel objects not present in the training set, while the kitchen environments remain the same. It evaluates the ability to adapt to novel objects using only a few demonstrations, \textit{e.g.}, learning the grasping strategy for the new object.
\\
\textit{L3 (Unseen Objects, Unseen Environment)}:  
This most challenging scenario requires the robot to perform manipulation tasks in entirely new kitchen environments with novel furniture layouts, backgrounds, and novel objects. This level thoroughly tests the robot's ability to generalize from a few demonstrations to completely novel scenarios, often requiring a completely novel motion sequence to solve the task, \textit{e.g.}, using a sink with a different faucet mechanism. 

\textbf{Embodiments.}
We consider two embodiments in simulation: a free-floating 6-DoF hand (Abstract) and a humanoid robot (GR1), both within the RoboCasa framework~\cite{robosuite,robocasa}.
This setup allows us to collect training data using the free-floating hand, while still evaluating the learned policy on the humanoid (GR1) platform, thereby simulating embodiment challenges~\cite{legato}.
It also enables a systematic analysis of the embodiment gap by comparing performance across the two embodiments.
In the real world, we collect human play videos and evaluate on the GR1 humanoid.
\begin{figure*}[ht!]
  \centering
  \begin{minipage}[t]{0.64\textwidth}
    \vspace{0pt} %
    \captionof{table}{Success rates in Abstract and GR1 embodiments in the simulation benchmark.}
    \vspace{5px}
    \resizebox{\linewidth}{!}{%
      \begin{tabular}{l ccc ccc}
      \toprule
      \multirow{2}{*}{\textbf{Method}} & \multicolumn{3}{c}{\textbf{Abstract}} & \multicolumn{3}{c}{\textbf{GR1}} \\
      \cmidrule{2-4} \cmidrule{5-7}
       & \textbf{L1} & \textbf{L2} & \textbf{L3} & \textbf{L1} & \textbf{L2} & \textbf{L3} \\
      \midrule
      H2R~\cite{h2r}                      & $0.03$ & $0.05$ & $0.03$ & $0.03$ & $0.00$ & $0.00$ \\
      Vid2Robot~\cite{vid2robot}          & $0.44$ & $0.40$ & $0.12$ & $0.41$ & $0.23$ & $0.11$ \\
      PEFT~\cite{hand}               & $0.47$ & $0.35$ & $0.00$ & $0.29$ & $0.21$ & $0.01$ \\
      \methodname{} w/o Visual Masking    & $0.59$ & $\mathbf{0.51}$ & $0.22$ & $0.37$ & $0.35$ & $0.09$ \\
      \methodname{}                       & $\mathbf{0.73}$ & $0.39$ & $\mathbf{0.27}$ & $\mathbf{0.59}$ & $\mathbf{0.44}$ & $\mathbf{0.26}$ \\
      \bottomrule
      \end{tabular}
    }
    \label{tab:performance_comparison}
  \end{minipage}
  \hfill
  \begin{minipage}[t]{0.35\textwidth}
    \vspace{0pt} %
    \centering
    \includegraphics[width=0.75\linewidth]{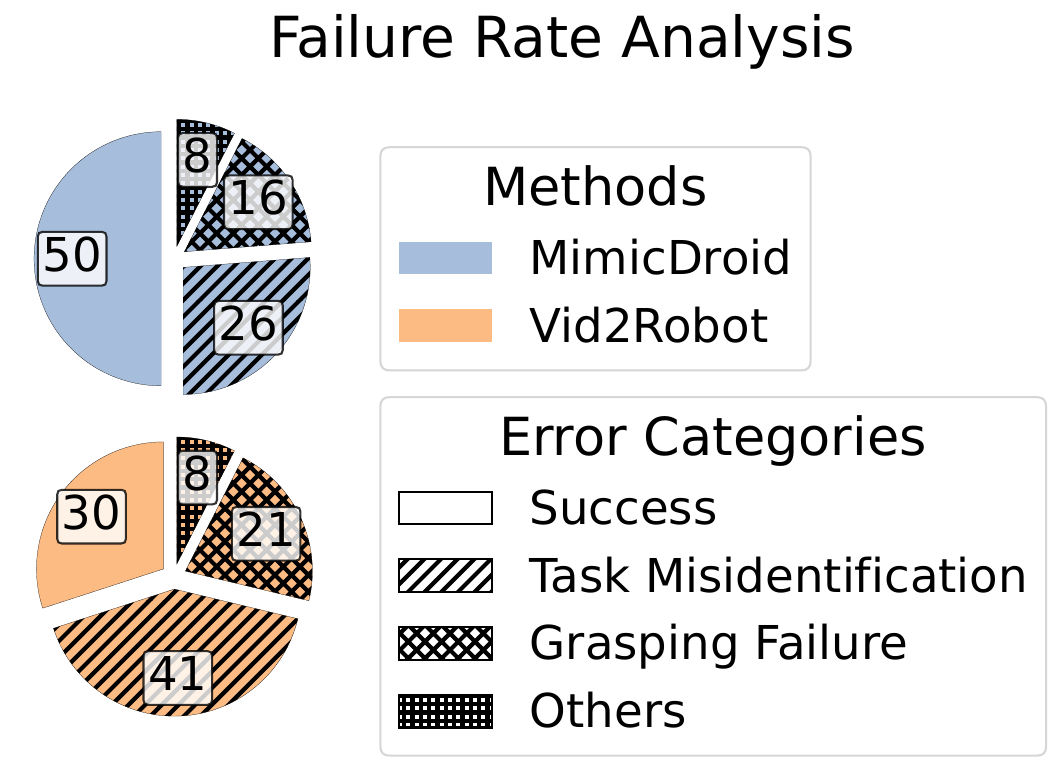}
    \captionof{figure}{\methodname{} reduces task misidentification and grasping errors using ICL compared to the video-conditioned baseline.}
    \label{fig:error_analysis}
  \end{minipage}
  \vspace{-10pt}
\end{figure*}

\textbf{Data-Collection Setup.}
We collect human play data by allowing an operator to interact with the scene freely. The operator performs meaningful tasks without specific goals, driven by curiosity.
For example, in a kitchen scenario, the operator might open the oven, put bread on the oven tray, and later move it to different receptacles. 
This free-form approach captures a richer diversity of interactions and object configurations than typical teleoperated demonstrations, since the operator is not constrained by specific task goals or task-specific resets.
Each play session lasts $20$ minutes in simulation and $10$ minutes in the real world. Interaction is done using a spacemouse in simulation and the operator’s hand in the real world.
We record each session as RGB videos in the real world, with randomized kitchen and object layouts.
We collect $8$ hours of simulated data ($320$k timesteps) and $45$ minutes of real-world data ($80$k frames).

\textbf{Implementation Details.}
During training, for each randomly sampled target trajectory segment, we find the top $k=10$ similar trajectories to serve as the context. The features for the trajectory are extracted using $f=\text{WiLoR}$~\cite{wilor} for hand pose estimation and $g=\text{DinoV2}$~\cite{dinov2} for visual features (Sec.~\ref{sec:method:training_sample}). In each iteration, we randomly mask patches of an image with probability $p=0.8$ and uniformly sample patches between $1$--$16$ (Sec.~\ref {sec:method:embodiment}). The model is trained to predict $l=32$ actions for each timestep (Sec.~\ref{sec:method:training_objective}). 

\textbf{Baselines.}
We compare~\methodname{} to the following baselines to evaluate its ability to perform ICL from human play videos.
(1) \textit{Task-conditioned methods.}
We use Vid2Robot~\cite{vid2robot}, which conditions on human videos, and H2R~\cite{h2r}, which conditions on final goal images.
Both lack action information in the context, a key component for modeling observation-action relationships for ICL.
This comparison isolates the effect of ICL, as these baselines cannot learn in-context.
(2) \textit{Fine-tuning methods.}
We compare~\methodname{} with the use of parameter-efficient fine-tuning (PEFT) to adapt at the test time for few-shot learning~\cite{hand}.
This comparison highlights the benefit of the instant, gradient-free adaptation of~\methodname{} via ICL over fine-tuning.
All baselines are trained on the same human play videos and use identical augmentations and training pipelines to avoid confounding factors and ensure fairness.

\section{Results}
In our evaluation, we aim to answer the following research questions and analyze \methodname{}'s failure cases.

\textbf{How well does~\methodname{} achieve generalization through ICL to downstream tasks?}
We compare~\methodname{} to image goal-conditioned H2R~\cite{h2r} and video-conditioned Vid2Robot~\cite{vid2robot} (Tab.~\ref{tab:performance_comparison}), two task-conditioned baselines that receive task specification to recognize the intended behavior, but lack the observation-action pairs needed to perform ICL.
In contrast, \methodname{} using ICL achieves performance gains of $+14\%$ and $+18\%$ in abstract and humanoid embodiments, respectively.
Furthermore, we compare with PEFT~\cite{hand} and find~\methodname{} not only learns instantly compared to test-time finetuning, but also achieves higher success rates ($+29\%$, $+26\%$).
Finetuning fails to adapt at the most challenging generalization level (L3), likely due to larger distribution shifts causing forgetting~\cite{yang2024selfdistillationbridgesdistributiongap,zhao2023does}, whereas ICL preserves pretrained knowledge due to gradient-free adaptation.
We also observe that test-time finetuning leads to overfitting on the abstract embodiment, evidenced by a larger performance drop from abstract to humanoid in PEFT ($-10\%$) compared to ICL ($-3\%$).

\begin{figure*}[ht!]
  \centering

  \begin{minipage}[t]{0.32\textwidth}
    \includegraphics[width=\linewidth]{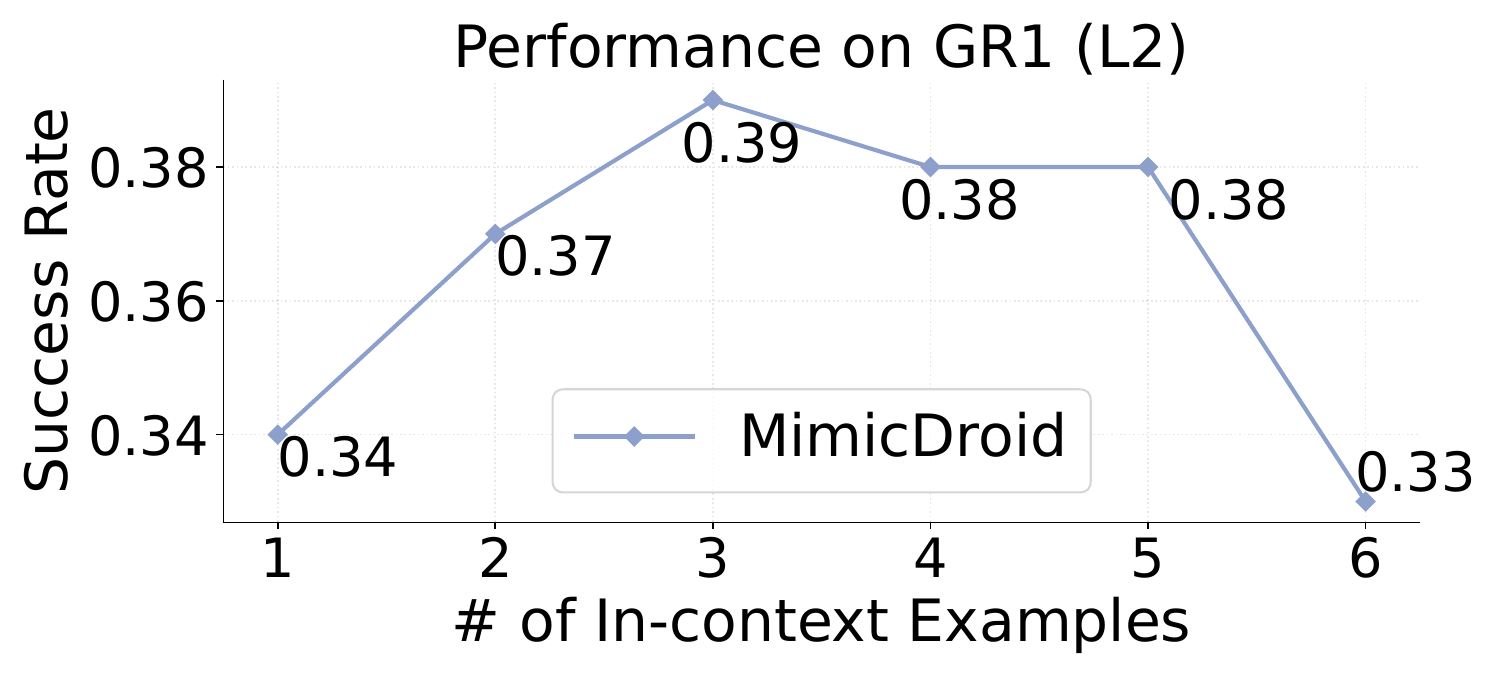}
    \captionof{figure}{Performance rises with more in-context examples but plateaus beyond 3 due to training-time context length.}
    \label{fig:wrap_n_prompts}
  \end{minipage}
  \hfill
  \begin{minipage}[t]{0.32\textwidth}
    \includegraphics[width=\linewidth]{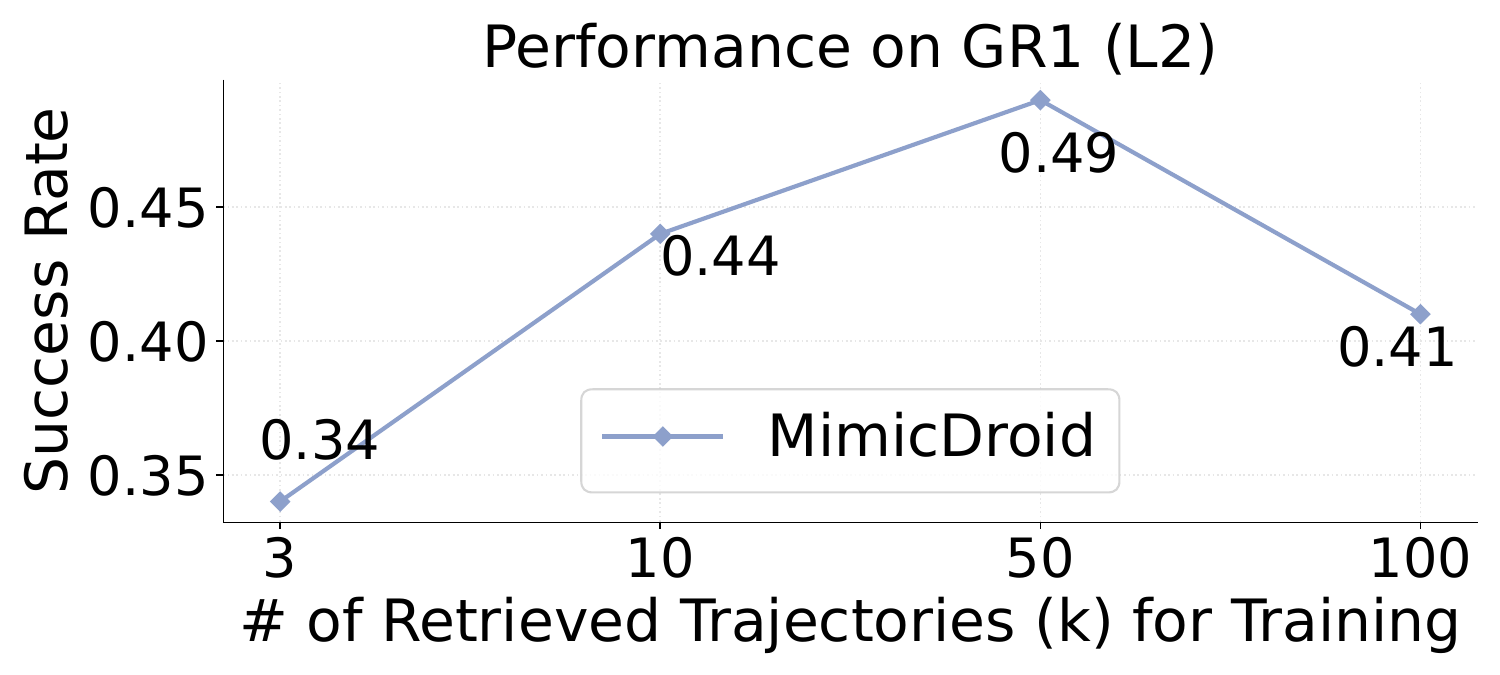}
    \captionof{figure}{Performance benefits from more retrieved context segments ($k$) per target, but high values introduce noise.} 
    \label{fig:wrap_meta_training}
  \end{minipage}
  \hfill
  \begin{minipage}[t]{0.32\textwidth}
    \includegraphics[width=0.95\linewidth]{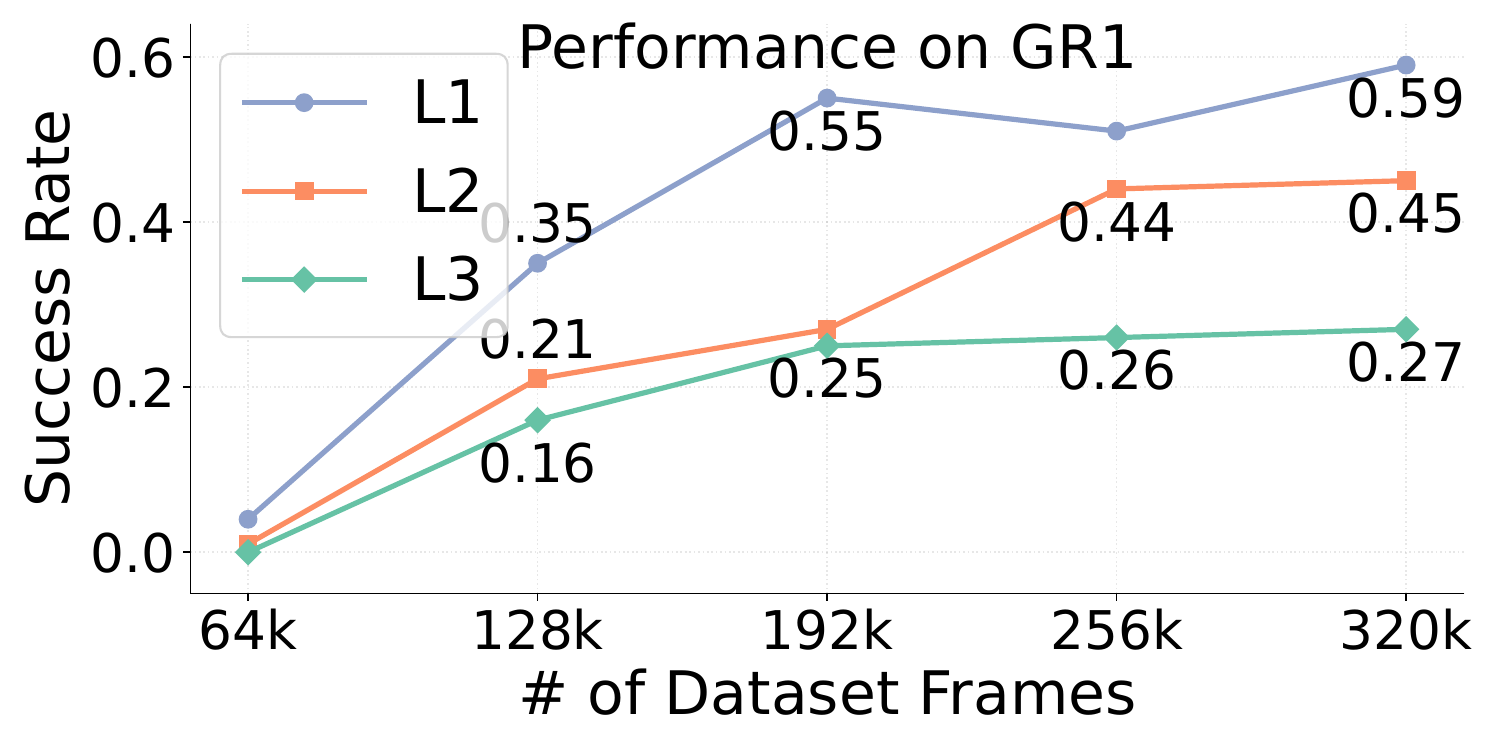}
    \captionof{figure}{Performance scales consistently with training data across from L1-L3, showing promise for learning from play.}
    \label{fig:scaling}
  \end{minipage}
\end{figure*}

\begin{figure*}[t!]
    \centering
    \includegraphics[trim={0 460px 480px 0}, width=0.9\textwidth]{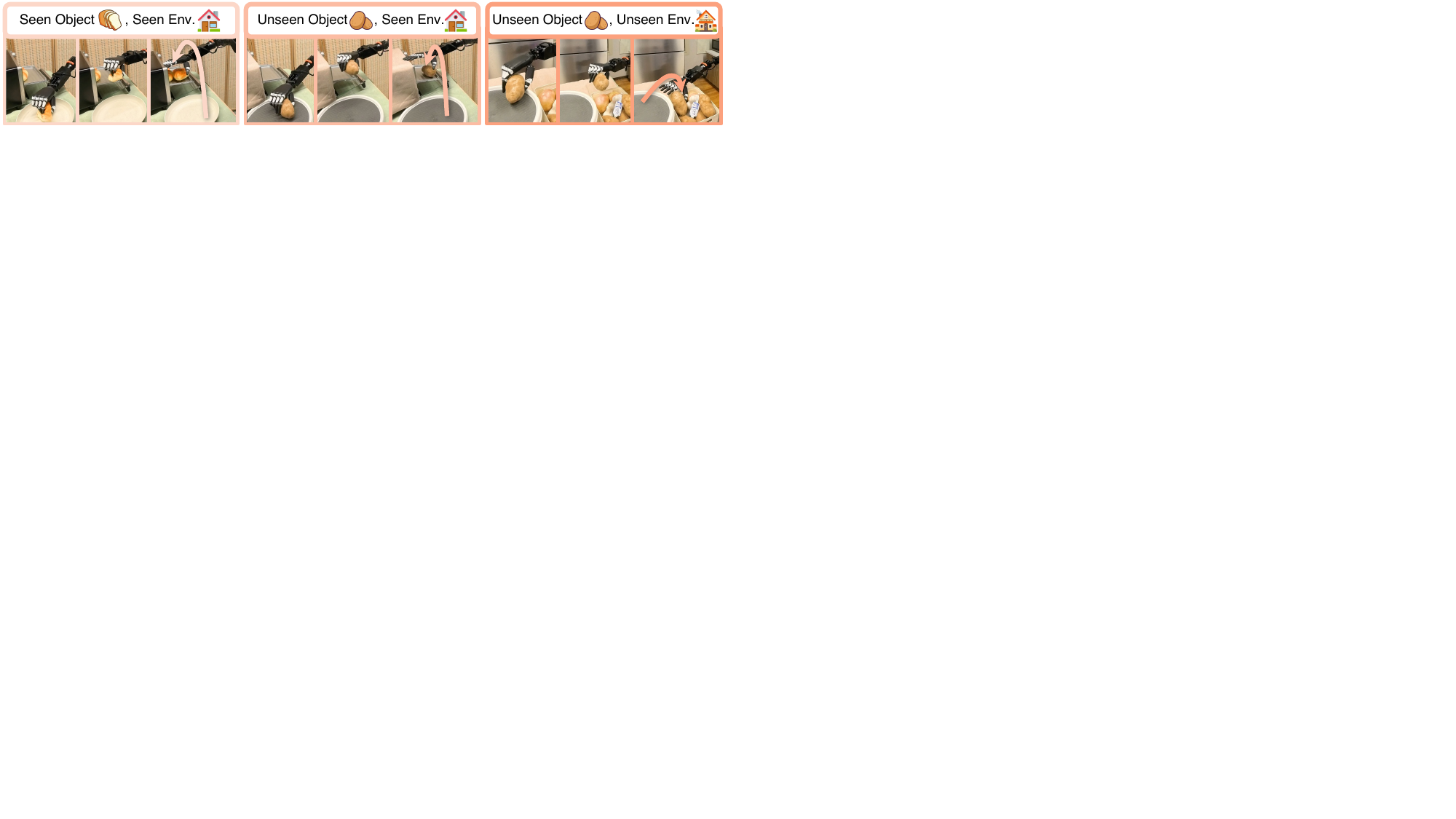}
    \caption{
    \textbf{Examples of real-world evaluations (L1-L3).}~\methodname{} generalizes to both seen (\textit{e.g.}, chips, bread, and apples) and unseen objects (\textit{e.g.}, potatoes, garlic, and cloth), as well as novel environments. Evaluation tasks include pick-and-place to different receptacles and articulated object manipulation (\textit{e.g.}, opening or closing an oven tray). 
    }
    \vspace{-10pt}
    \label{fig:rw_results}
\end{figure*}
\textbf{Real-world results.} Furthermore, we evaluate~\methodname{} on the GR1 humanoid in the real world, using a policy trained solely on human play videos.~\methodname{} achieves a success rate of $\mathbf{0.53}$ in L1, $\mathbf{0.23}$ in L2, $\mathbf{0.08}$ in L3, nearly twofold higher than Vid2Robot, which obtains $0.28$, $0.08$, $0.00$, respectively.
These results highlight~\methodname{}’s ability to adapt to novel objects and scenes using few demonstrations via ICL (Fig.~\ref{fig:rw_results}).

\textbf{How effective is~\methodname{} at bridging the visual gap between embodiments?}
We ablate visual masking to assess its role in bridging the visual gap between embodiments.
In the last two rows of Table~\ref{tab:performance_comparison}, removing masking (\methodname{} w/o Visual Masking) causes a sharp drop in transfer performance to the humanoid ($-17\%$) compared to only $-3\%$ with~\methodname{}, averaged over all three levels. Although it performs better in source L2, this is likely due to overfitting to the abstract embodiment.
Moreover, we evaluate an alternative strategy, EgoMimic~\cite{kareer2024egomimic}, which masks the hands with a black patch and red line. In the real world, \methodname{}'s random patch masking during training ($0.53\%$) achieves comparable performance to EgoMimic ($0.58\%$) while eliminating the need for external modules such as SAM~\cite{sam} for segmentation.
These results highlight the effectiveness of random patch masking for robustness to the visual gap between embodiments.

\textbf{How well~\methodname{}'s performance scales with the number of in-context examples?}
We evaluate how \methodname{}'s performance varies with the number of context demonstrations provided at test time. In Fig.~\ref{fig:wrap_n_prompts}, performance improves as the number of context examples increases from $1$ to $3$. However, the gains drop with $4$ to $6$ examples due to the limited context length supported by the transformer policy during training.
Naively extending the context length is expensive as both memory and compute cost scale linearly with sequence length; future work can explore efficient ways to handle more in-context examples.

\textbf{How does the quality of training data used in the Meta-ICL affect ICL performance?}
To evaluate the impact of data quality on Meta-ICL, we vary the number of similar context segments retrieved during training, $k$. As shown in Fig.~\ref{fig:wrap_meta_training}, performance drops for large $k$ due to the inclusion of dissimilar segments, while small $k$ limits training diversity. 
We find that $k=50$ strikes a balance between maintaining training diversity and avoiding excessive noise from dissimilar segments.
These results highlight the importance of selecting context-target pairs with meaningful observation-action similarity for effective Meta-ICL.
Future work can explore more robust data curation strategies for Meta-ICL.

\textbf{How does scaling dataset size impact~\methodname{}'s ability to perform ICL?}
We observe consistent performance improvements across all generalization levels (L1-L3) as the amount of training data increases, demonstrating the scalability of learning from RGB play videos (Fig.~\ref{fig:scaling}).
Specifically, L1 improves from $35\%$ at $128$k frames to $59\%$ at $320$k, and L2 rises from $21\%$ to $45\%$, resulting in a $+24\%$ absolute gain for both.
L3 also improves, from $16\%$ to $27\%$ ($+11\%$), though the gains are less pronounced compared to L1 and L2.
These results affirm the benefits of scaling training data, while also motivating a more systematic study of the factors that influence ICL performance on harder generalization tasks like L3.

\textbf{Failure Analysis.}
Failures in downstream tasks arise from task misidentification ($26\%$), missed grasps ($16\%$), and other errors ($8\%$) like incomplete cabinet closure, missed placement (Fig.~\ref{fig:error_analysis}).
Compared to Vid2Robot,
\methodname{} notably reduces both misidentification ($-15\%$) and grasping errors ($-5\%$) using ICL.

Despite these gains,~\methodname{} still struggles with few-shot performance in L3 (Tab.~\ref{tab:performance_comparison}). It struggles to learn tasks requiring novel robot motions, potentially due to higher learning complexity. For instance, faucet activation in seen environments involves left-to-right motions, whereas L3 environments introduce sinks that require novel, bottom-to-top motions. 
Lastly, in real-world settings,~\methodname{} overfits to motions of specific hand sizes seen during training, leading to collisions in cluttered environments.
We hypothesize that this issue can be mitigated by using data from various operators, which we leave to future work.

\section{Limitations and Future Work}
We aim to improve~\methodname{} by addressing several limitations.
First, it relies on human play videos, which provide high-quality videos for learning to learn in-context. A natural extension is to augment with web-scale human videos (\textit{e.g.}, YouTube) to expand object and environment diversity.
Second,~\methodname{} extracts actions via an off-the-shelf hand pose predictor. While these models handle partial occlusions with hand priors, they fail when hands vanish (\textit{e.g.}, reaching into cupboards or behind furniture). Combining hand and full-body motion estimation~\cite{slahmr} may help.
Finally,~\methodname{} treats demonstrations as low-level state-action sequences, learning how but not why motions matter. Thus, it cannot generalize across semantically equivalent tasks. Meta-training with language-trajectory pairs, akin to Flamingo~\cite{flamingo}, may enable this capability.
\section{Conclusion}
We introduce \methodname{}, an in-context learning method for few-shot learning in humanoid manipulation.
When deployed,~\methodname{} infers humanoid action from a few human videos, possibly involving novel objects and environments via in-context learning.
\methodname{} acquires this capability by learning from continuous human play videos, leveraging a scalable and diverse data source. 
It achieves this by leveraging the similar manipulation behaviors in human play data as a self-supervised signal to meta-train the policy for in-context learning.
To bridge the visual and kinematic embodiment gaps between humans and humanoids, ~\methodname{} uses random patch masking to reduce overfitting to human appearances and retargets human hand poses to humanoid wrist poses to preserve task intent.
To support systematic evaluation, we introduce a new simulation benchmark for assessing few-shot learning in humanoid manipulation.
Our results, both in simulation and the real world, highlight the promise of leveraging in-context learning from human play videos, with a notable twofold improvement in performance.
In conclusion, this work introduces a new method for learning in-context learning from diverse and scalable human play videos, laying the groundwork for future research towards adaptive robot assistants for everyday environments.

\section*{Acknowledgments}
We thank Ben Abbatematteo, Haonan Chen, and Bowen Jiang for providing valuable feedback on the manuscript. 
We also thank all members of the Robot Perception and Learning Lab and the Robot Interactive Intelligence Lab at UT Austin for their insightful discussions.
This work was partially supported by the National Science Foundation (FRR-2145283, EFRI-2318065), the Office of Naval Research (N00014-24-1-2550), the DARPA TIAMAT program (HR0011-24-9-0428), and the Army Research Lab (W911NF-25-1-0065). It was also supported by the Institute of Information \& Communications Technology Planning \& Evaluation (IITP) grant funded by the Korean Government (MSIT) (No. RS-2024-00457882, National AI Research Lab Project).

\printbibliography

\appendix
\section{Appendix}
\subsection{Data Collection}
We collect human play data in both real-world and simulated settings to train our in-context learning policy.  
In the real world, a human operator interacts freely with the environment using a single hand for approximately $10$ minutes per play session.  
These sessions are recorded using a single static RGB camera.  
In simulation, the operator teleoperates two free-floating hands using a $6$-DoF space mouse for $20$-minute sessions.  
Each simulated session includes both the teleoperated action trajectories and corresponding RGB observations from three cameras, which are necessary to cover the larger workspace.
Table~\ref{tab:data-collection} summarizes the statistics.
\begin{table}[h]
\centering
\caption{Human Play Data Collection Summary}
\label{tab:data-collection}
\begin{tabular}{lccc}
\toprule
\textbf{Modality} & \textbf{Duration per Session} & \textbf{Total Hours} & \textbf{Total Frames} \\
\midrule
Real world        & $10$ minutes                   & $1$ hour               & $80,000$ \\
Simulation        & $20$ minutes                   & $8$ hours              & $320,000$ \\
\bottomrule
\end{tabular}
\end{table}

\subsection{Data Preprocessing}\label{app:preprocessing}
To prepare the raw play data for training, we convert it into structured (observation, action) pairs.
\textbf{Real world:} Hand poses are estimated per frame using the WiLoR model~\cite{wilor}. The action at timestep $t$ is defined as the hand pose at $t+1$ ($k=1$), consisting of the 6-DoF wrist pose (wrt workspace camera) and a binary gripper signal indicating open or closed.
\textbf{Simulation:} Action information (absolute wrt robot base frame) is obtained from teleoperation

This procedure allows us to create processed trajectories $\tau' = \{s_t', a_t\}$ where $s_t'$ consists of the object-centric crops from RGB frames and proprioceptive hand pose, and $a_t$ is the target action.
\subsection{Model Architecture}
Our policy model is built on a long-context transformer backbone with modality-specific encoders.  
Training samples consist of $3$ context trajectories and $1$ target trajectory.  
Each trajectory is $256$ steps long in simulation and $128$ in the real world to mirror the sufficiently large horizon length of a single task in each domain.
Each step includes both observation and action tokens, but to reduce computational cost, input frequency is downsampled by a factor of $4$.
Note the output action frequency still remains the same as a $32$ chunk of actions are predicted at every step.
As a result, the effective context length during training becomes:
\begin{itemize}
    \item Simulation: $4\text{ (\# of Traj.)} \times 64$ (observations) + $4\text{ (\# of Traj.)} \times 64$ (actions) = $512$ tokens
    \item Real world: $4\text{ (\# of Traj.)} \times 32$ (observations) + $4\text{ (\# of Traj.)} \times 32$ (actions) = $256$ tokens
\end{itemize}

\vspace{0.5em}
\noindent
\textbf{Model Components:}
We built upon the existing open-source codebase of ICRT~\cite{icrt}.
\begin{itemize}
    \item \textbf{Observation Encoder:} The observation encoder processes both proprioceptive and object-centric observations.
    \begin{itemize}
        \item \textbf{Proprioception Encoder:} A $2$-layer MLP maps a $16$-/$8$- dimensional proprioceptive input (\textit{e.g.}, wrist pose + gripper information) to a $768$-dimensional embedding.
        \item \textbf{Vision Encoder:} A pretrained model Cross-MAE~\cite{crossmae} is used to extract visual features. This pretrained model is frozen during training.
        \item \textbf{Visual Attention Fusion:} The proprioceptive embeddings are combined with visual features using a self-attention module. This fusion allows the model to ground the proprioceptive state in the visual context before being passed to the policy.
    \end{itemize}
    
    \item \textbf{Action Encoder:} A $2$-layer MLP encodes the $14$-/$7$-dimensional action vector ($6$-DoF pose + gripper state) into a $768$-dimensional embedding.

    \item \textbf{Transformer Backbone:} A $4$-layer LLaMA-style transformer processes the sequence of observation and action tokens. It uses RMSNorm, rotary positional embeddings (RoPE) in the transformer block to support long-context reasoning across trajectories.

    \item \textbf{Action Decoder:} An MLP head decodes the transformer output corresponding to a state into a $32\times\text{action-dim}$-dimensional prediction vector representing a chunk of $32$ future actions (each of dimension $14$/$7$). During rollouts, only the first $16$ actions from the chunk are executed before predicting the action again.
\end{itemize}

\begin{table}[h]
\centering
\caption{Model Summary}
\label{tab:model-arch}
\resizebox{0.99\columnwidth}{!}{%
\begin{tabular}{ll}
\toprule
\textbf{Component} & \textbf{Details} \\
\midrule
Vision Encoder     & Pre-trained CrossMAE~\cite{crossmae} \\
Proprio Encoder    & MLP: $16/8 \rightarrow 768$ \\
Visual Attention Pooling  & $3$ layers, $768 \rightarrow 256$ \\
Action Encoder     & MLP: $14/7 \rightarrow 768$ \\
Transformer        & $4$ layers, LLaMA-style, RMSNorm, $768$-dim \\
Action Decoder     & MLP head, $768 \rightarrow 32\times14/32\times7$ \\
\bottomrule
\end{tabular}
}
\end{table}
\subsection{Training Details}
\begin{table}[h]
\centering
\caption{Training Hyperparameters}
\label{tab:training-config}
\resizebox{0.99\columnwidth}{!}{%
\begin{tabular}{ll}
\toprule
\textbf{Config} & \textbf{Value} \\
\midrule
Optimizer & AdamW \\
Base Learning Rate & $5$e-$4$ \\
Effective Batch Size & 256 \\
Weight Decay & $0.01$ \\
Warmup Epochs & $2$ \\
Total Epochs & $200$ \\
Sequence Length & $512$ (Simulation), $256$ (Real world) \\
Num Action Prediction Steps & $32$ \\
Proprioception Noise & $0.005$ \\
Brightness Augmentation & Uniform($-0.1$, $0.1$) \\
Contrast Augmentation & Uniform($0.8$, $1.2$) \\
Random Patch Masking & $0$–$16$ patches, Area $\sim$ Uniform($1\%$, $4\%$) \\
Num Cameras & $3$ (Simulation), $1$ (Real world) \\
\bottomrule
\end{tabular}
}
\end{table}
Training is performed over 200 epochs using $8\times \text{A}5000$ GPUs, completing in approximately two days.  
Each RGB frame is first augmented with random brightness, contrast, and up to $16$ erased patches covering $1\%$–$4\%$ of the image area.  
Refer to Table~\ref{tab:training-config}.
\subsection{Real-world Evaluation}
Models are trained on the human play videos and evaluated on the humanoid (GR1) embodiment. For each task, we take $10$ rollouts on the robot.
We structure evaluation into three levels of increasing difficulty:
\begin{itemize}
    \item \textbf{L1 (Seen Objects, Seen Environment):} Tasks use the same objects and environment as training.
    \begin{itemize}
        \item Pick and Place from Oven-top to Plate
        \item Pick and Place from Rotating Disk to Plate
        \item Pick and Place from Plate to Oven Tray
        \item Pull Out Oven Tray
    \end{itemize}
    \item \textbf{L2 (Unseen Objects, Seen Environment):} Tasks involve novel objects from unseen categories 
    (potato, garlic, cloth) as well as new instances from seen categories (oven, bowl).
    \begin{itemize}
        \item Pick and Place from Oven-top to Bowl
        \item Pick and Place from Rotating Disk to Plate
        \item Pick and Place from Plate to Oven Tray
        \item Pull Out Oven Tray
    \end{itemize}
    \item \textbf{L3 (Unseen Objects, Unseen Environment):} Tasks are evaluated in a new environment using 
    novel objects, where novel objects are similar to those in L2.
    \begin{itemize}
        \item Pick and Place from Oven-top to Bowl
        \item Pick and Place from Drawers to Plate
        \item Pick and Place from Plate to Oven Tray
        \item Pull Out Oven Tray
    \end{itemize}
\end{itemize}

\subsection{Simulation Benchmarking}
Our simulation benchmark is designed to systematically evaluate the few-shot generalization capabilities of humanoid manipulation. We build upon the \texttt{robocasa} framework with a diverse set of environments ($8$ in training) and objects ($30$ in training). Models are trained on the play data collected from abstract (free-floating hand) and evaluated on both abstract (free-floating hand) and humanoid (GR1) embodiment. For few-shot learning, prompts are drawn from a held-out validation set of play data in abstract embodiment that represent the target task, and provided to the model at evaluation time.

We evaluate on three levels of generalization (Sec.~\ref {sec:exp_setup_main}).
Each generalization level includes four distinct tasks, as shown in Table~\ref{tab:simulation_all}.
L1 tasks involve seen objects and environments from training, but with new object positions.
L2 tasks introduce novel objects to be manipulated, \textit{e.g.}, an onion, garlic, or a new cabinet.
L3 tasks feature novel objects in a new environment. This level is particularly difficult because it often requires novel robot motions; for example, the new sink may have a different base height than in training environments, the faucet mechanism may operate top-to-bottom rather than left-to-right, or the task may involve moving an object between receptacles with completely different heights.

\begin{table*}[t]
\centering
\caption{Evaluation is structured into three levels with increasing difficulty and 4 tasks in each level.}
\renewcommand{\arraystretch}{1.2}
\begin{tabular}{|c|c|c|c|}
\hline
\textbf{Level} & \textbf{Task Name} & \textbf{Abstract Embodiment} & \textbf{Humanoid Embodiment} \\
\hline
\multirow{4}{*}{L1 (Seen Objects, Seen Environment)} 
 & PnPSinkToRightCounterPlate & \includegraphics[width=0.22\linewidth]{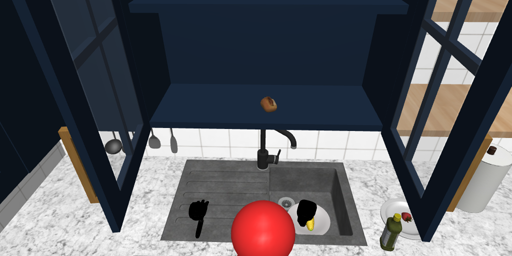} & \includegraphics[width=0.22\linewidth]{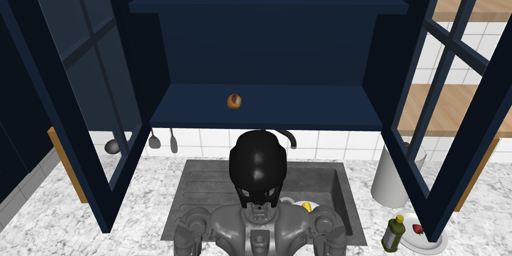} \\
 & PnPSinkToCabinet & \includegraphics[width=0.22\linewidth]{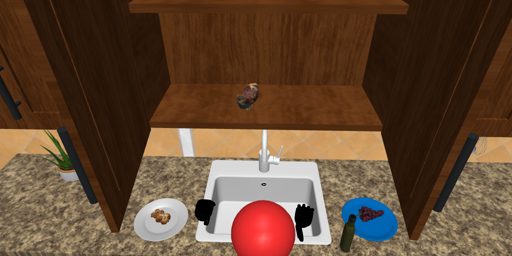} & \includegraphics[width=0.22\linewidth]{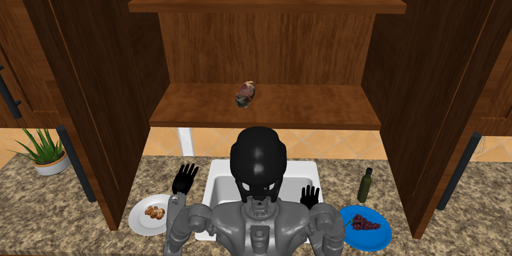} \\
 & TurnOnFaucet & \includegraphics[width=0.22\linewidth]{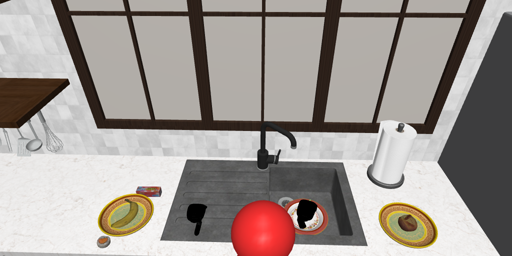} & \includegraphics[width=0.22\linewidth]{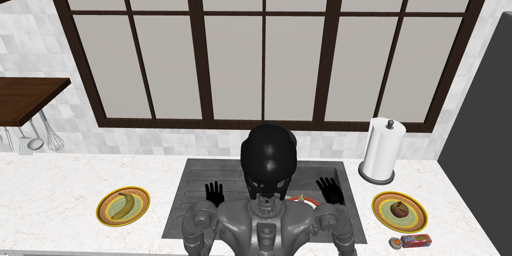} \\
 & CloseLeftCabinetDoor & \includegraphics[width=0.22\linewidth]{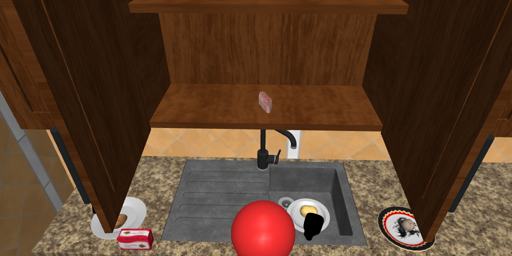} & \includegraphics[width=0.22\linewidth]{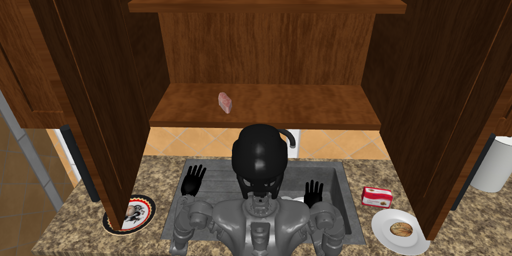} \\
\hline
\multirow{4}{*}{L2 (Unseen Objects, Seen Environment)} 
 & PnPSinkToRightCounterPlateL2 & \includegraphics[width=0.22\linewidth]{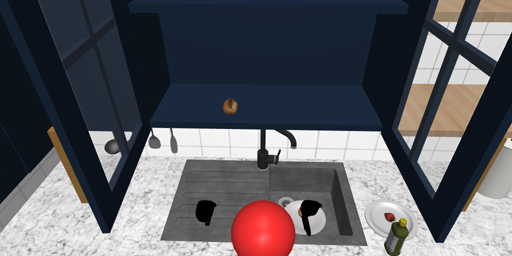} & \includegraphics[width=0.22\linewidth]{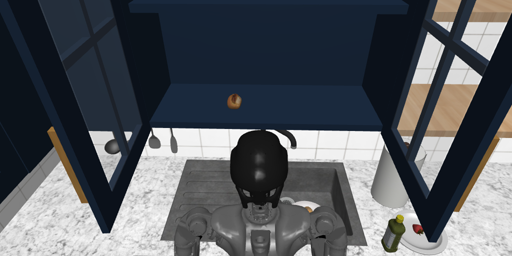} \\
 & PnPSinkToCabinetL2 & \includegraphics[width=0.22\linewidth]{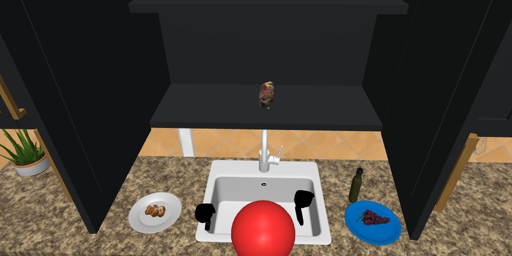} & \includegraphics[width=0.22\linewidth]{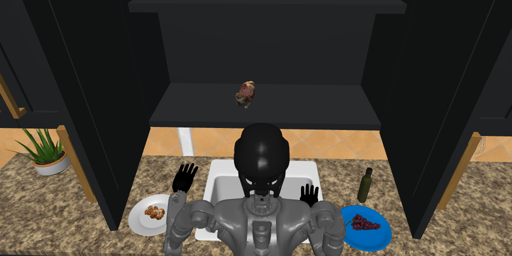} \\
 & CloseRightCabinetDoorL2 & \includegraphics[width=0.22\linewidth]{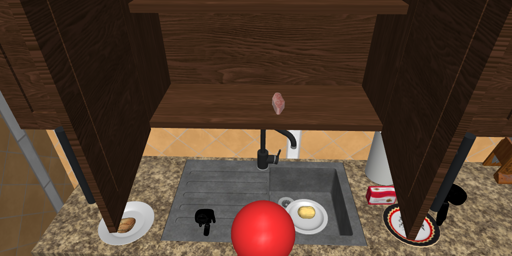} & \includegraphics[width=0.22\linewidth]{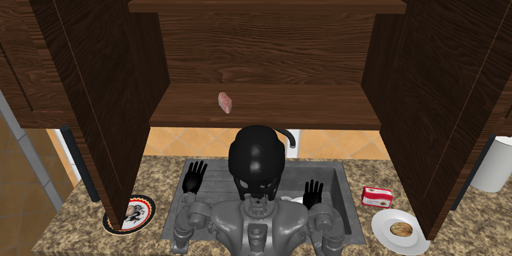} \\
 & CloseLeftCabinetDoorL2 & \includegraphics[width=0.22\linewidth]{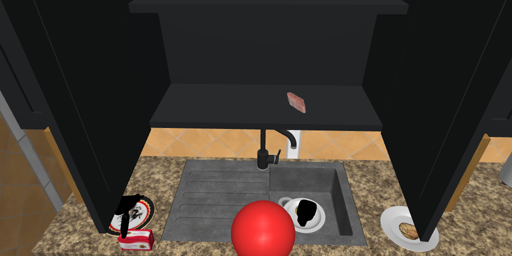} & \includegraphics[width=0.22\linewidth]{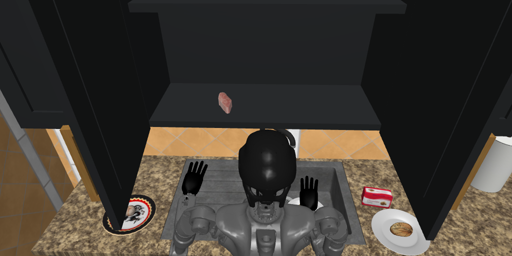} \\
\hline
\multirow{4}{*}{L3 (Unseen Objects, Unseen Environment)} 
 & CloseLeftCabinetDoorL3 & \includegraphics[width=0.22\linewidth]{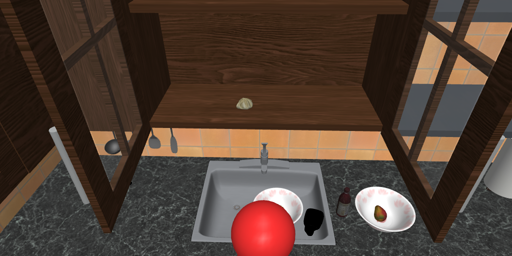} & \includegraphics[width=0.22\linewidth]{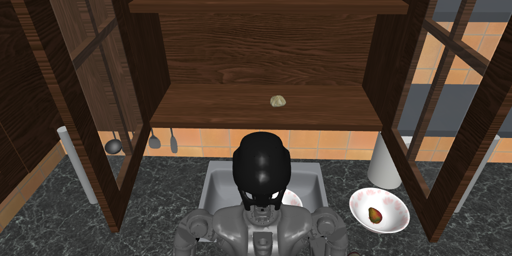} \\
 & PnPSinkToRightCounterPlateL3 & \includegraphics[width=0.22\linewidth]{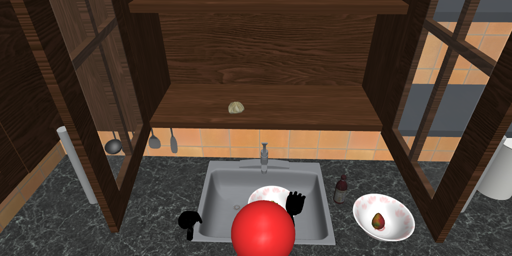} & \includegraphics[width=0.22\linewidth]{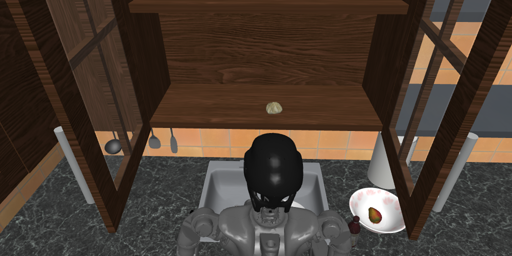} \\
 & PnPSinkToMicrowaveTopL3 & \includegraphics[width=0.22\linewidth]{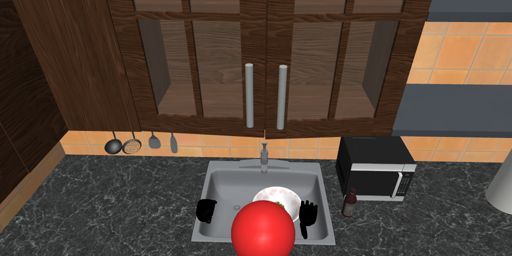} & \includegraphics[width=0.22\linewidth]{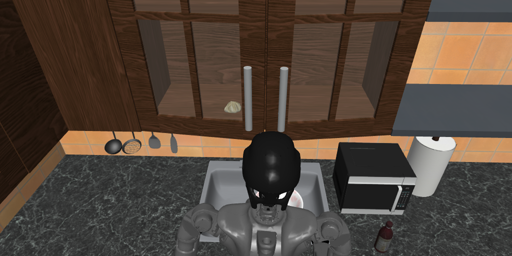} \\
 & TurnOnFaucetL3 & \includegraphics[width=0.22\linewidth]{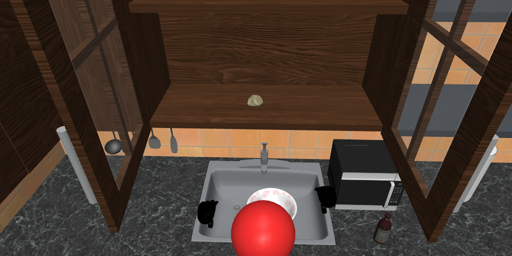} & \includegraphics[width=0.22\linewidth]{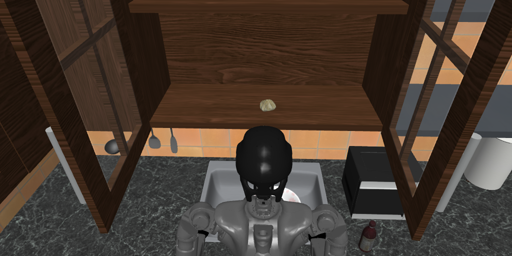} \\
\hline
\end{tabular}
\label{tab:simulation_all}
\end{table*}

\newpage
\clearpage

\end{document}